\documentclass[10pt,twocolumn,letterpaper]{article}

\usepackage[pagenumbers]{cvpr} 
\usepackage{pifont}
\usepackage{multirow}
\usepackage{booktabs}
\usepackage[dvipsnames]{xcolor}

\definecolor{cvprblue}{rgb}{0.21,0.49,0.74}

\usepackage{hyperref}
\hypersetup{colorlinks,allcolors=black}

\def\paperID{7746} 
\def\confName{CVPR}
\def\confYear{2024}

\title{In-Context Matting}

\author{He Guo\\
\and
Zixuan Ye\\
\and
Zhiguo Cao\\
\and
Hao Lu \thanks{Corresponding author}\\
\and
School of Artificial Intelligence and Automation,\\
Huazhong University of Science and Technology, China\\
{\tt\small\{hguo01,hlu\}@hust.edu.cn}
}

\begin{document}

\twocolumn[{%
\renewcommand\twocolumn[1][]{#1}%
\maketitle

\begin{center}
\vspace{-1pt}
\setlength{\abovecaptionskip}{3pt}
\setlength{\belowcaptionskip}{5pt}
\captionsetup{type= float type}
\small
\centering
\setlength{\tabcolsep}{1pt}
\renewcommand{\arraystretch}{0.7}

\includegraphics[width=\linewidth]{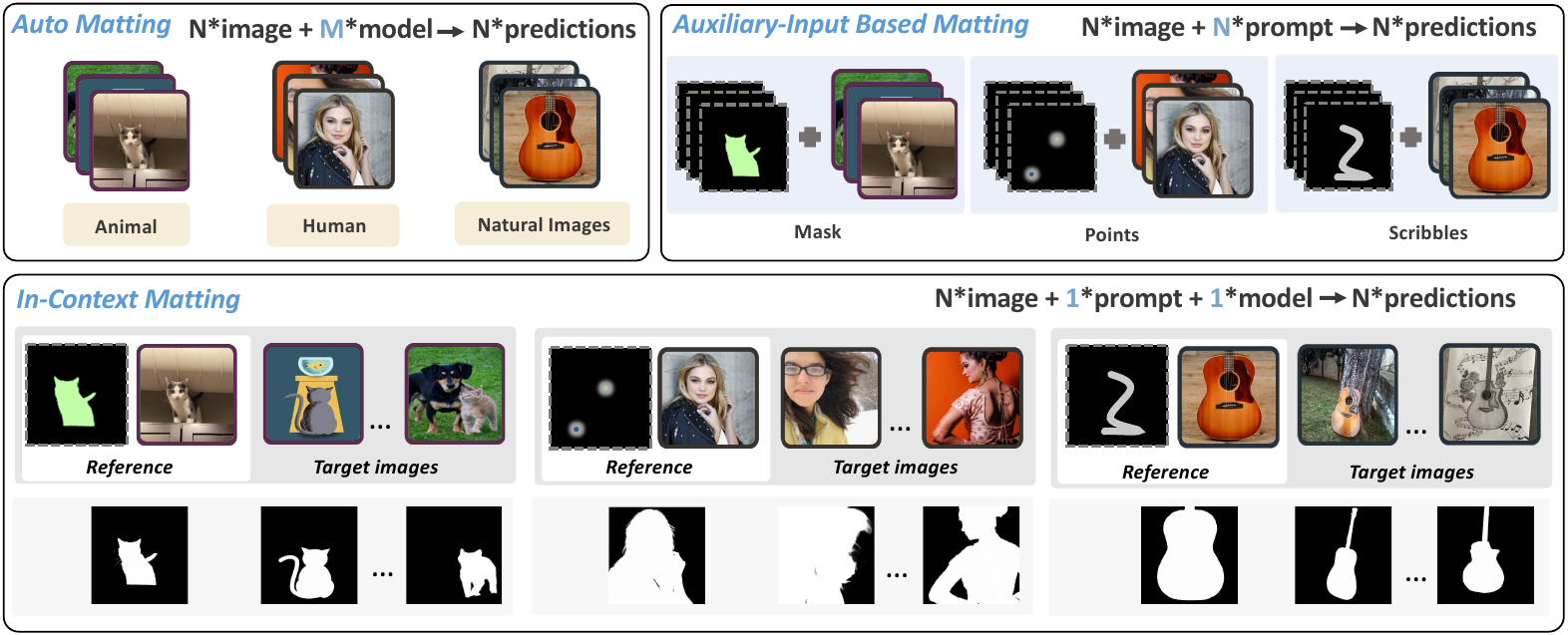}
\captionof{figure}{
\textbf{In-Context Matting.}
This novel task setting for image matting enables automatic natural image matting of target images of a certain object category conditioned on a reference image of the same category, with user-provided priors such as masks and scribbles on the reference image only. Notice that, our approach exhibits remarkable cross-domain matting quality.
}
\label{fig:fig1}
\end{center}
}]

\renewcommand{\thefootnote}{}
\footnotetext{*Corresponding author}

\begin{abstract}
We introduce in-context matting, a novel task setting of image matting. Given a reference image of a certain foreground and guided priors such as points, scribbles, and masks, in-context matting enables automatic alpha estimation on a batch of target images of the same foreground category, without additional auxiliary input. 
This setting marries good performance in auxiliary input-based matting and ease of use in automatic matting, which finds a good trade-off between customization and automation. 
To overcome the key challenge of accurate foreground matching, we introduce IconMatting, an in-context matting model built upon a pre-trained text-to-image diffusion model. Conditioned on inter- and intra-similarity matching, IconMatting can make full use of reference context to generate accurate target alpha mattes. To benchmark the task, we also introduce a novel testing dataset ICM-$57$, covering $57$ groups of real-world images.
Quantitative and qualitative results on the ICM-$57$ testing set show that IconMatting rivals the accuracy of trimap-based matting while retaining the automation level akin to automatic matting. Code is available at \url{https://github.com/tiny-smart/in-context-matting}.

\end{abstract}

\section{Introduction}
\label{sec:intro}

Image matting has been a long-standing problem in vision and graphics~\cite{boda2018survey}. It typically requires estimating an accurate alpha matte by solving a so-called matting equation
\begin{equation}
    I=\alpha F+(1-\alpha) B\,,
\end{equation}
where $I$ is the $3$-channel RGB image, $F$, $B$, and $\alpha$ are the $3$-channel foreground, the $3$-channel background, and the $1$-channel alpha matte.
The matting equation, however, is highly ill-posed, due to the need to infer $7$ unknowns from $3$ observations.

Prior art has come up with different ways to reduce uncertainties in matting such as using trimaps~\cite{DIM,index,timi,GCA,LBSampling,CA,lfp}, scribbles~\cite{yang2022unified}, or a even known background~\cite{BM1,BM2}.
These approaches are given the name auxiliary input-based matting~\cite{li2023deep} in modern matting literature. Indeed, these matting models, particularly trimap-based ones, have achieved remarkable accuracy. They are in some way user-unfriendly as an auxiliary input should be provided with each image in practice, which significantly harms matting efficiency and user experience.
Recently another stream of work attempts to abandon any auxiliary input completely and forms a new paradigm called automatic matting~\cite{ma2023rethinking,li2021deep,sun2022human,li2021privacy,li2022bridging,chen2018semantic,lf}.
Despite their inherent advantages, these automatic matting models are narrowed to specific object categories, such as humans~\cite{ma2023rethinking,sun2022human,li2021privacy,chen2018semantic}, animals~\cite{li2022bridging}, and salient objects~\cite{li2021deep,lf}. This can render poor generalization in natural scenes, infeasibility to tackle general object categories, and unawareness of foreground of interest.

Hence, there seems an obvious gap between accuracy and efficiency and between customization and automation. An interesting question is that:
\textit{Can the auxiliary input-based matting be optimized to enhance the efficiency, while also maintaining guidance for matting targets with sufficient automation, thereby harmonizing the two existing matting paradigms?}

In this paper, we introduce \textit{in-context matting}, a novel task setting of image matting, where a reference input can provide guidance for a batch of target images with similar foregrounds. The alpha mattes of the target batch are predicted by leveraging the contextual information from the reference input. 
Fig.~\ref{fig:fig1} provides an example, where a cat marked in the reference image enables the extraction of cats from all target images, regardless of the background or 
domain.
This novel task setting relieves users from providing auxiliary input for each image. Instead, by specifying the matting target only in a single reference image, the alpha mattes of the entire batch images could be predicted. 
Given the reference guidance, in-context matting can also gather 
sufficient contextual information, leading to higher accuracy and adaptability than fully automatic matting. This setting therefore combines the most notable features of automatic and auxiliary input-based matting, finding a good trade-off between them.

Technically, we confront a primary challenge inherent to in-context matting, \ie, how to 
leverage the reference input to accurately identify the corresponding target foreground. While SegGPT~\cite{wang2023seggpt} has explored image segmentation with contextual information, its in-context coloring framework is not suitable for image matting. 
In this work, we approach this challenge as a problem of region-to-region matching. 
In particular, recent advances~\cite{song2019generative,nichol2021glide,rombach2022high} in generative diffusion models have demonstrated 
emergent capabilities in discriminative tasks like segmentation~\cite{xu2023open} and correspondence~\cite{tang2023emergent}. Given that our region-to-region matching 
is an inherent aspect of correspondence, and noting the parallels between image segmentation and image matting, we explore the 
applicability of pretrained diffusion models for in-context matting.

We therefore introduce IconMatting, a model based on the pre-trained 
Stable Diffusion model~\cite{rombach2022high}, 
specialized for in-context matting. 
Given a reference image along with its corresponding 
foreground of interest as context, 
the target alpha matte could be inferred by exploiting the feature correspondence 
from Stable Diffusion such that the target foreground is matched conditioned on the correspondence. 
However, 
the matching is often sparse and insufficient to represent the entire 
target area. 
To address this, the intra-image similarity, based on the self-attention maps of Stable Diffusion, is additionally used to supplement the missing parts. By leveraging both inter- and intra-image similarities, 
informative guidance 
of the matting target would be acquired. 
Finally, any off-the-shelf 
matting heads 
can be used to predict the alpha matte.



Since the task setting is different from existing matting benchmarks, we introduce a new testing dataset named ICM-$57$ to offer a broad and thorough validation of in-context matting. This dataset encompasses $57$ contextually-aligned image groups; each comprising images in the real world and has either the same category or the same instance of different views, thereby encompassing a rich variety of in-context scenarios, which ensures a comprehensive test of a model to tackle various 
context.

Through extensive experiments on the ICM-$57$ and AIM-$500$~\cite{AIM} datasets, we showcase the potential of in-context matting and IconMatting. The results 
indicate that IconMatting, while retaining the automation level akin to automatic matting, rivals the accuracy of trimap-based matting, underscoring the value of in-context matting as a promising direction for image matting. 

Our 
contributions include: 
\begin{itemize}
\item We introduce in-context matting, a novel task setting of image matting that takes advantages of both automatic and auxiliary input-based matting;

\item IconMatting: 
an effective in-context matting model based on Stable Diffusion.

\item ICM-$57$: 
a novel dataset and the evaluation framework for in-context matting. 
\end{itemize}


\section{Related Work}
\label{sec:related_work}
We review work related to image matting and in-context learning in vision.
\vspace{-10pt}
\paragraph{Image Matting.}
Image matting approaches can be 
coarsely categorized into auxiliary input-based matting and automatic matting.
Auxiliary input-based matting requires user input. 
The user input can be in the form of a trimap~\cite{DIM,timi,GCA,LBSampling,CA,lfp}, scribbles~\cite{yang2022unified}, a background image~\cite{BM1,BM2}, a coarse mask~\cite{MGM,MGMwild}, or even a text description~\cite{li2023referring}. 
Despite their effectiveness, they require significant manual effort to provide the auxiliary inputs.
Automatic matting~\cite{HATT,AIM,chen2018semantic,li2022bridging, lf} 
predicts the alpha matte without any user intervention. They typically assume 
salient or certain foregrounds that are implicitly defined by the training dataset. The network structures used in automatic matting can be divided into two groups: one-stage network with global guidance~\cite{HATT} and parallel multi-task network~\cite{sun2022human,AIM}. Some recent work has also introduced transformer structures into automatic matting~\cite{ma2023rethinking}.
While 
both auxiliary input-based and automatic matting have been studied comprehensively, a paradigm that combines the efficiency of automatic matting and the precision of auxiliary input-based matting has not yet been explored. 
We fill this gap with in-context matting.

\vspace{-10pt}
\paragraph{In-Context Learning in Vision.}

In-context learning, initially a concept in natural language processing~\cite{dong2022survey}, is now made popular in computer vision. It allows models to 
fast adapt to a variety of tasks with minimal examples.

Bar \etal~\cite{bar2022visual} first proposes an in-context learning framework using inpainting with discrete tokens on figures and infographics from vision articles, demonstrating 
applications in foreground segmentation, single object detection, and colorization. Painter~\cite{wang2023images} adopts masked image modeling 
to perform in-context learning with supervised datasets, 
achieving highly competitive results on seven diverse tasks.
More recently, SegGPT~\cite{wang2023seggpt}, which segments everything in context by unifying different segmentation tasks into an in-context coloring framework. Prompt Diffusion~\cite{gong2023prompting} presents a diffusion-based generative framework to enable in-context learning across various tasks. Additionally, Flamingo~\cite{alayrac2022flamingo}, a family of visual language models, shows rapid adaptation to a variety of image and video tasks with few-shot learning capabilities. These models showcase the potential of in-context learning in addressing diverse 
vision tasks.

The concept of in-context learning, while being transformative in other areas, has not yet 
impacted the field of image matting. Although SegGPT achieves in-context image segmentation but is limited to coarse levels, lacking in semi-transparency handling. Our IconMatting 
first introduces in-context learning into image matting, enhancing both the efficiency of auxiliary input-based matting and the precision of automatic matting.

Our work is also related to image co-segmentation~\cite{rother2006cosegmentation}. This task aims to segment common objects within a pair of contextual images. However, unlike in-context matting, where users can specify the matting target, co-segmentation operates without user input, predicting rough binary masks delineating common objects across image pairs.


\section{In-Context Matting with Diffusion Models}
We begin with the problem setup, then present our proposed in-context matting model, \ie, IconMatting.

\subsection{Problem Setup}
\label{sec:problem_setup}
The objective of in-context matting is to extract the alpha mattes $\{ {\alpha _i}\} ^N$ of a specified foreground category from a collection of input images $\{ {I_i}\} ^N$. Through user interaction, the matting target is indicated by a reference image \( I_r \) and a corresponding binary region of interest (RoI) map \( M_{\tt RoI} \). The RoI map can take the form of a mask, scribbles, or points as exemplified in Fig.~\ref{fig:fig1}. Notably, the reference image can either be a part of the input collection or an entirely separate image. When the input image collection has only a single image, users can treat that image as the reference image. In this case, in-context matting degenerates into image matting guided by user interaction.

Given $\{ {I_i}\} ^N$ and \( (I_r, M_{\tt RoI}) \), in-context matting is formulated as predicting the alpha matte $\{ {\alpha _i}\} ^N$ in $\{ {I_i}\} ^N$ of the matting targets informed by \( (I_r, M_{\tt RoI}) \). In the context of in-context matting, when provided with a reference input, it becomes an automatic matting system targeted towards a specific foreground. The comparison between in-context matting and existing task settings for image matting is detailed in Table~\ref{tab:ts}.

\begin{table}[!t]
\footnotesize
\renewcommand{\arraystretch}{1}
\addtolength{\tabcolsep}{2pt}
\begin{tabular}{lcccc}
\hline
\multicolumn{1}{c}{Task setting} & auto & \begin{tabular}[c]{@{}c@{}}auto with\\ ref-input\end{tabular} & \begin{tabular}[c]{@{}c@{}}real-world\\ generalization\end{tabular} & \begin{tabular}[c]{@{}c@{}} \#ref-input\end{tabular} \\ \hline
Aux    & \ding{55}    & \ding{55}                   & good                 & \#images                    \\
Auto                & \ding{51}    & -                 & poor                 & zero                    \\
In-context               & \ding{55}    & \ding{51}                   & good                 & one                    \\ \hline
\end{tabular}
\caption{\textbf{Comparison between in-context matting and two existing image matting paradigms.} ``Aux'' and ``Auto'' are abbreviations for auxiliary input-based matting and automatic matting, respectively. In-context matting 
requires only a single reference input to achieve the automation of automatic matting and the generalizability of auxiliary input-based matting.}
\label{tab:ts}
\end{table}

\subsection{Overall Architecture}
\begin{figure*}[!t]
  \centering
  \includegraphics[width=\linewidth]{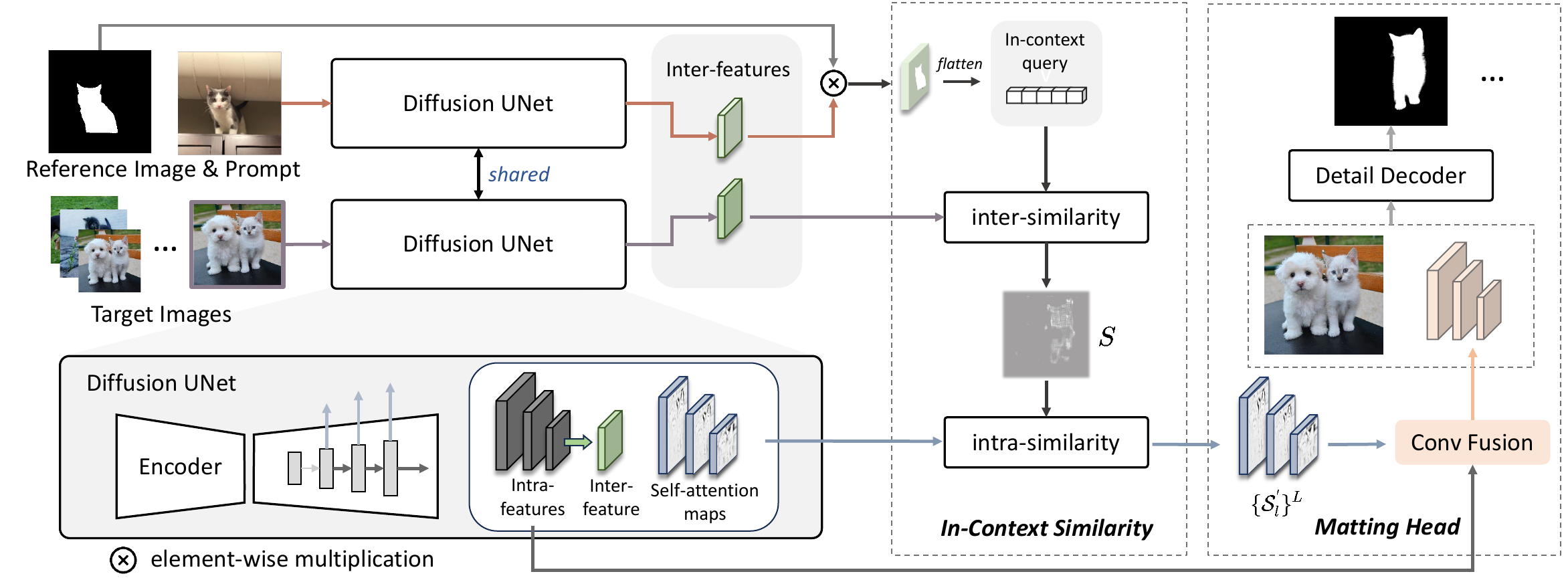}
  \vspace{-10pt}
  \caption{\textbf{IconMatting} integrates a Stable Diffusion-derived feature extractor, an in-context similarity module, and a matting head. It processes a target image \( I_t \), a reference image \( I_r \), and an RoI map \( M_{RoI} \).  Both reference and target image features and target self-attention maps are extracted and used. In-context similarity uses the in-context query from the reference image to create a guidance map, which, combined with self-attention maps, assists in locating the target object. The matting head finally generates the target alpha matte.}
  \label{fig:pipeline}
\end{figure*}





Here we first present the overall framework of IconMatting.
As shown in Fig.~\ref{fig:pipeline}, IconMatting is comprised of three 
components: a feature extractor, an in-context similarity module, and a matting head.

The feature extractor is responsible for obtaining the RoI features from the reference image, the features and self-attention maps from the target image. 
They are then fed into the in-context similarity module, the core of the framework. The module further consists of inter- and intra-similarity sub-modules: the former leverages the reference RoI features as an in-context query to derive a guidance map from the target features; the latter integrates the guidance map with multi-scale self-attention maps to obtain guidance for the matting head.
Finally, the matting head uses this synthesized guiding information and the target features to generate the alpha matte of the target image.

\label{sec:method}
\subsection{In-Context Feature Extractor}
\label{Preliminary}

\paragraph{Backbone Selection.}

As outlined in Section~\ref{sec:intro}, we conceptualize the core challenge of in-context matting -- leveraging the reference context to accurately identify the target foreground -- as a region-to-region matching problem. Therefore, if the features derived from a backbone naturally possess correspondence capabilities, referred to as in-context features, they would facilitate the implementation of in-context matting. Tang \etal~\cite{tang2023emergent} have found that the text-to-image generative model, Stable Diffusion~\cite{rombach2022high}, trained on large-scale text-image paired datasets, exhibit emergenent capabilities for both geometric and semantic correspondence. It can 
perform point-to-point correspondence between images across instances, classes, and even domains with simple cosine similarity. Inspired by this observation, we leverage Stable Diffusion as a feature extractor to implement in-context matting.


\vspace{-10pt}
\paragraph{Preliminary on Stable Diffusion.}

Recent advances in diffusion models, \eg, Stable Diffusion~\cite{rombach2022high}, have shown impressive results in both generative and discriminative tasks. 
Being a feature extractor, Stable Diffusion encodes an image \( x_0 \) into a latent space, denoted by \( z_0 \), which, through a noise process defined by \( \{\alpha_t\}^T \), transforms into $z_t = \sqrt{\alpha_t} z_0 + (\sqrt{1 - \alpha_t}) \boldsymbol{\varepsilon}$, where 
\( \boldsymbol{\varepsilon} \sim \mathcal{N}(0, \mathbf{I}) \) is the randomly-sampled noise. The latent representation \( z_t \) undergoes forward propagation in a U-Net \( f_\theta \), generating multi-scale features ${\{ {{\cal F}_l}\} ^L}$ and self-attention maps ${\{ {{\cal A}_l}\} ^L}$, which can later be exploited for downstream tasks. IconMatting uses the capabilities of Stable Diffusion and both reference and target images to extract multi-scale features and self-attention maps to enhance feature representation.

\subsection{In-Context Similarity}
In-context similarity plays a key role in our model, because the quality of inferred alpha mattes highly depends on the output of this module.
In particular, the in-context similarity module is designed to identify the potential target foreground taking the reference RoI into account, thereby guiding the prediction of the target alpha matte. 
According to our 
observations, both the reference-target similarity and target-target similarity matter for locating the potential target foreground. These correspond to the proposed inter-similarity and intra-similarity sub-modules.

\vspace{-10pt}
\paragraph{Observation.}
\begin{figure}[htbp]
  \centering
  \includegraphics[width=\linewidth]{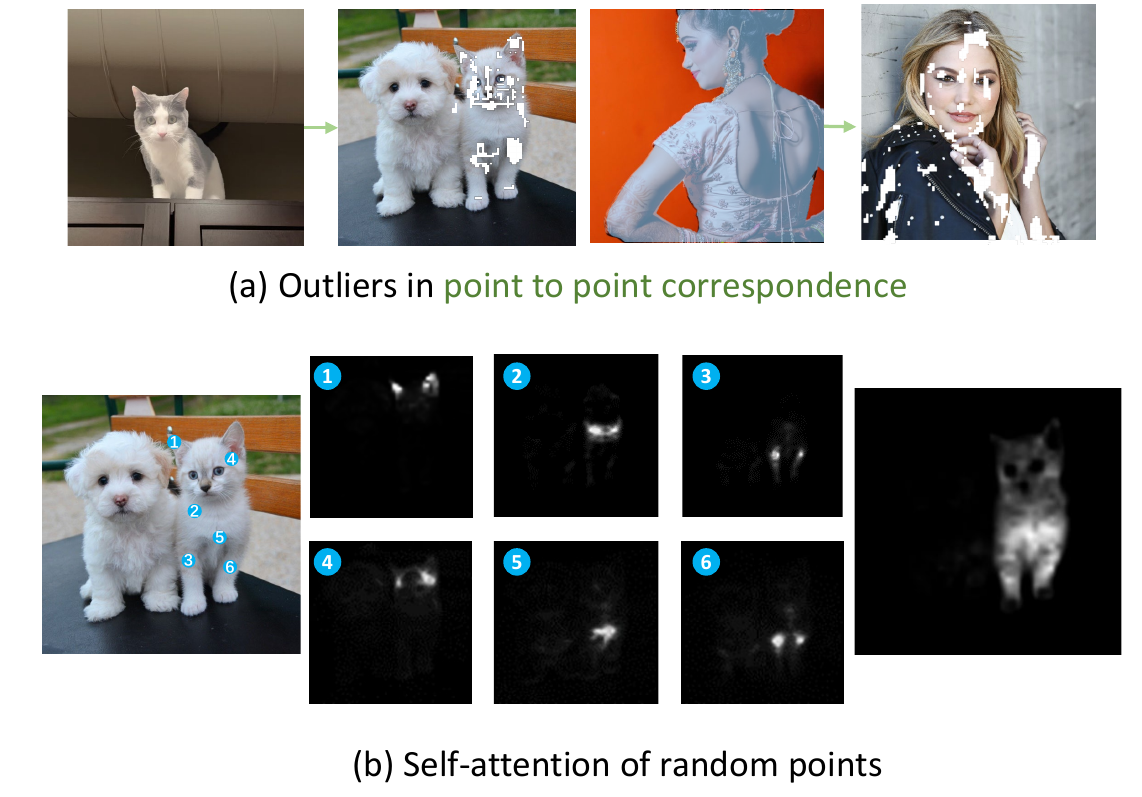}
  \vspace{-20pt}
  \caption{\textbf{Observations on the inter- and intra similarities.}}
  \label{fig:ob}
\end{figure}

The core challenge in in-context matting is a semantic correspondence problem. 
Given Stable Diffusion being the feature extractor, one can 
associate points of the foreground areas between the reference and target images using the emergent feature correspondence. However, due to the inherent reference-target foreground difference, a 
rigorous one-to-one mapping of all points between the two areas is unfeasible. There would always be some unmatched outliers, resulting in holes in the matting target area of the target image, as illustrated in Fig.~\ref{fig:ob}. 

Since only a subset of points 
are matched, the goal is changed to how to expand these matched points to cover the entire target foreground area. To address this, we look for other points sharing similar semantic meaning with this subset of points. Intra-image similarity is therefore considered. Intuitively, the self-attention maps from Stable Diffusion reflect the similarities between different image patches. As shown in Fig.~\ref{fig:ob}, by randomly sampling a small number of points from the target area and simply summing over the corresponding self-attention maps, the whole target foreground can be revealed. Based on this observation, we use the self-attention maps as intra-image information to supplement the inter-image matching results.

\vspace{-10pt}
\paragraph{Inter-Similarity.}




Formally, given features ${\{ {{\cal F}_l^r}\} ^L}$ and ${\{ {{\cal F}_l^t}\} ^L}$ extracted from the reference image $I_r$ and the target image $I_t$, respectively, the layer with the best correspondence capability is selected following DIFT~\cite{tang2023emergent}, 
denoted by ${\cal F}_{\tt inter}^r$ for the reference and ${\cal F}_{\tt inter}^t$ for the target. Then, features corresponding to the RoI $ M_{\tt RoI} $ in ${\cal F}_{\tt inter}^r$ are extracted and formulated as the in-context query ${\left\{ {{Q_k}} \right\}^K}$, where $K$ is the length of the query. ${Q_k}$ takes the form
\begin{equation}
    {Q_k} = R\left( k \right)\,,
\end{equation}
\begin{equation}
    R = {\cal F}_{\tt inter}^r \odot {M_{\tt RoI}}\,,
\end{equation}
where $\odot$ is the element-wise product and $R\left( k \right)$ denotes the $k$-th non-zero element of $R$.
Further, ${\left\{ {{Q_k}} \right\}^K}$ is used to compute similarity with ${\cal F}_{\tt inter}^t$, identifying regions on \( I_t \) that 
correspond to the RoI of \( I_r \), formulated as
\begin{equation}
    {{\cal S}_k} = {\tt softmax}(\frac{{{Q_k} \cdot F{{_{\tt inter}^t}^T }}}{{\sqrt d }})\,.
\end{equation}
The similarity map is denoted by \( \{ S_k \} ^K \), and the mean of all such similarity maps yields \( S \), which measures the degree of similarity between different locations on \( I_t \) and the RoI in \( I_r \), serving as the first intermediate output of in-context similarity.

\vspace{-10pt}
\paragraph{Intra-Similarity.}
As noted in our observations (Fig.~\ref{fig:ob}), \( S \) is typically sparse. Although the target matting region on \( I_t \) is partially covered by \( S \), it is insufficient to 
guide alpha prediction. Here we further design the intra-similarity sub-module to leverage the internal similarity within \( I_t \) to propagate \( S \) into a more precise representation of the RoI on \( I_t \). During feature extraction of \( I_t \), self-attention maps ${\{ {{\cal A}_l}\} ^L}$ representing its internal similarity are also retained, serving as the input for intra-similarity. This sub-module uses \( S \) as a weight to the self-attention maps, thereby generating guiding information that accurately represents the matting target on \( I_t \), 
denoted by \( \{ {\cal S}_l^{'} \}^{L} \). Mathematically, the intra-similarity matching is expressed by
\begin{equation}
    {\cal S}_l^{'} = {{\cal A}_l} \odot {\cal S}\,.
\end{equation}

\subsection{Matting Head}

The success of ViTMatte~\cite{vitmatte} implies that 
the information of original image is important during decoding. Following this practice, in our matting head, the original image is concatenated and decoded with outputs from previous modules. 

The guidance map from the in-context similarity module and the intra-features from the backbone are merged and refined using a convolutional feature fusion block, including a series of convolution, normalization, and activation layers. The output multi-scale in-context features are progressively merged using a series of fusion layers which comprise upsampling, concatenation, convolution, normalization, and activation layers. Then, following ViTMatte~\cite{vitmatte}, details from the original image are extracted and combined with the merged feature 
in a detail decoder, enhancing the details of alpha matte. This matting head effectively melds contextual information with original image details, yielding the generation of a highly precise and refined alpha matte.

\subsection{Reference-Prompt Extension}
In addition to the mask prompts, points and scribbles can also be transformed into RoI masks. However, these prompts yield less comprehensive in-context queries compared with mask prompts. To enhance them, we propose an extension of reference prompt to enrich the in-context queries derived from point and scribble prompts.

Since the self-attention maps from the backbone reflect the similarities across regions, we leverage the attention maps of the reference images to expand the RoI mask. This is achieved by including regions in the attention maps that are similar to the prompt locations. Specifically, for each prompt point, the top $m$ points with the highest responses in their corresponding attention maps are integrated into the RoI mask additionally, thus enriching the in-context query.

\section{Results and Discussion}

%





\subsection{Datasets}
To facilitate 
in-context matting, we establish a hybrid training set, 
along with a test set, ICM-$57$. The existing AIM-$500$ dataset was also reorganized to meet the testing requirements of in-context matting. In particular, 
in-context matting 
requires to organize images into groups where the annotated foregrounds share categories or instances. Such organization allows for random selection of reference and target images within groups during training. In the test set, one or more images in each group are designated as reference images. 

\vspace{-10pt}
\paragraph{Mixing In-Context Training Sets.}

We selected the RM-1K dataset~\cite{10198480}. 
However, such a dataset 
was insufficient for 
training. Therefore, a subset of the Open Images dataset~\cite{kuznetsova2020open}, focusing on image segmentation, was also employed. The two formed a mixed training set tailored to in-context matting. Both datasets were reorganized. 

For the RM-1K dataset, we divided it into $222$ groups. 
A subset of $14,000$ images from the Open Images dataset were chosen as well. 
In the original dataset, each image came with one or more annotations for image segmentation; each corresponds to an object instance. We aggregated these annotations by category, 
ensuring that the annotations include all instances of the corresponding category. 
They subsequently formed context groups that 
met the requirements of in-context matting. As a result, a mixed training set 
of $15,000$ images and $450$ groups was created.

\begin{figure}[!t]
  \centering
  \includegraphics[width=\linewidth]{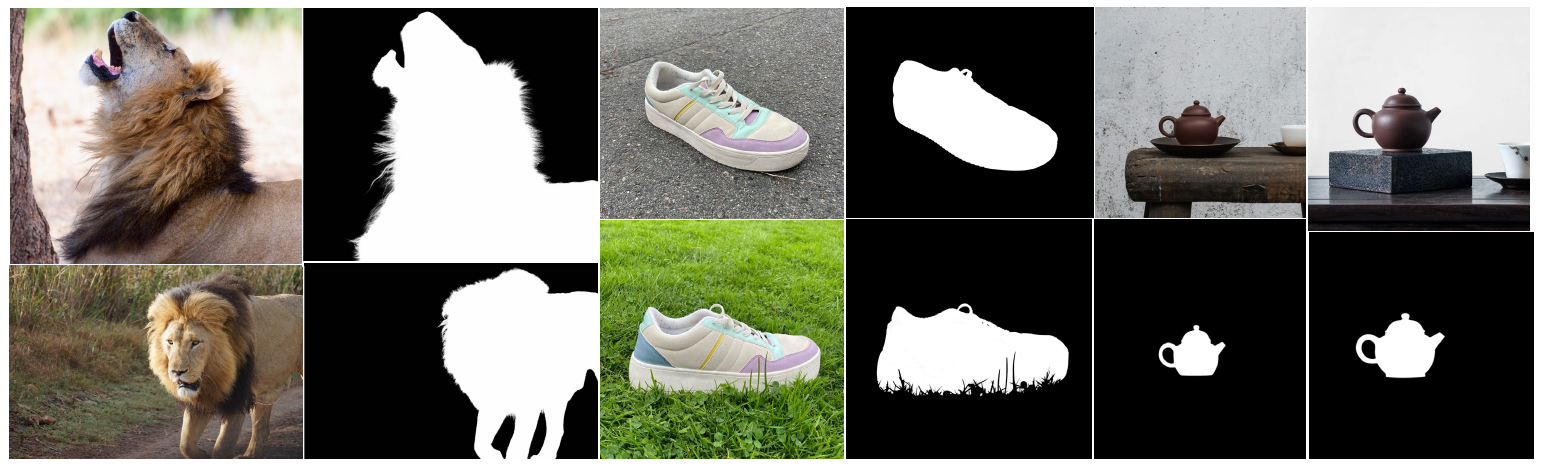}
  \caption{\textbf{ICM-$57$ examples}. The dataset encompasses foreground subjects including human, animals, plants, and various common objects. It contains both instances from the same category and the same entity.}
  \label{fig:dataset}
\end{figure}

\vspace{-10pt}
\paragraph{ICM-57 Testing Set.}
To assess the performance of our model, we constructed the first testing dataset for in-context matting, named ICM-$57$, which comprises $57$ image groups that form various real-world  contexts. This dataset was created by using the instance-segmented dreambooth dataset~\cite{ruiz2022dreambooth}, to which we supplemented high-precision alpha matte annotations. Additionally, we reviewed the existing AIM-$500$ dataset, selected a subset, and categorized these into groups according to their classes to supplement the in-category groups. Examples of the testing set are shown in Fig~\ref{fig:dataset}.

\vspace{-10pt}
\paragraph{AIM-500.} We also report performance on the AIM-500 dataset~\cite{AIM}. This enables us to compare our matting model with other public matting results on this dataset.

\subsection{Implementation Details}

\paragraph{Architecture.}
We employ the U-Net architecture from Stable Diffusion v2-1-v~\cite{rombach2022high}. 
For feature extraction, the time step of the diffusion process is set to $t = 0$ by default, with an empty string used for conditional input. U-Net has $11$ decoder blocks; we extract feature maps from the $5$-th, $8$-th, and $11$-th blocks as the intra-features and ones from the $5$-th block 
as the inter-features.

\vspace{-10pt}
\paragraph{Training Details.}
We employ distinct loss functions for matting and segmentation, respectively. For matting, we use a combination of $\ell_1$ loss, Laplacian loss, and Gradient loss. For segmentation, we only use the $\ell_1$ loss.
To leverage the segmentation dataset while reducing the impact of imprecise edge annotations, we adopt the approach from HIM~\cite{sun2022human} that only 
backpropagates the loss from the confident areas. 

During training, the learning rate is set to $0.0004$ and the batch size is $8$. The input images are randomly cropped to a size of $768\times768$ pixels. To prevent deviation from the pre-trained model space in modeling real images, no additional data augmentation is used. 
We train IconMatting for $20,000$ iterations using the AdamW optimizer.

\vspace{-10pt}
\paragraph{Evaluation.}

We employ the four widely used matting metrics: SAD, MSE, Grad and Conn~\cite{metrics}. 
Lower values imply higher-quality mattes. 
In particular, MSE is scaled by a factor of \(1 \times 10^{-3}\).

To reduce randomness, 
each method 
is tested for three rounds, where the metrics were 
averaged. 
For each group of images, the 
reference inputs 
were fixed and consistently used. 
This 
minimizes the variations in reference inputs, 
allowing for a more scientific and reliable assessment.

\subsection{Main Results}
\begin{figure*}[!t]
  \centering
  \includegraphics[width=\linewidth]{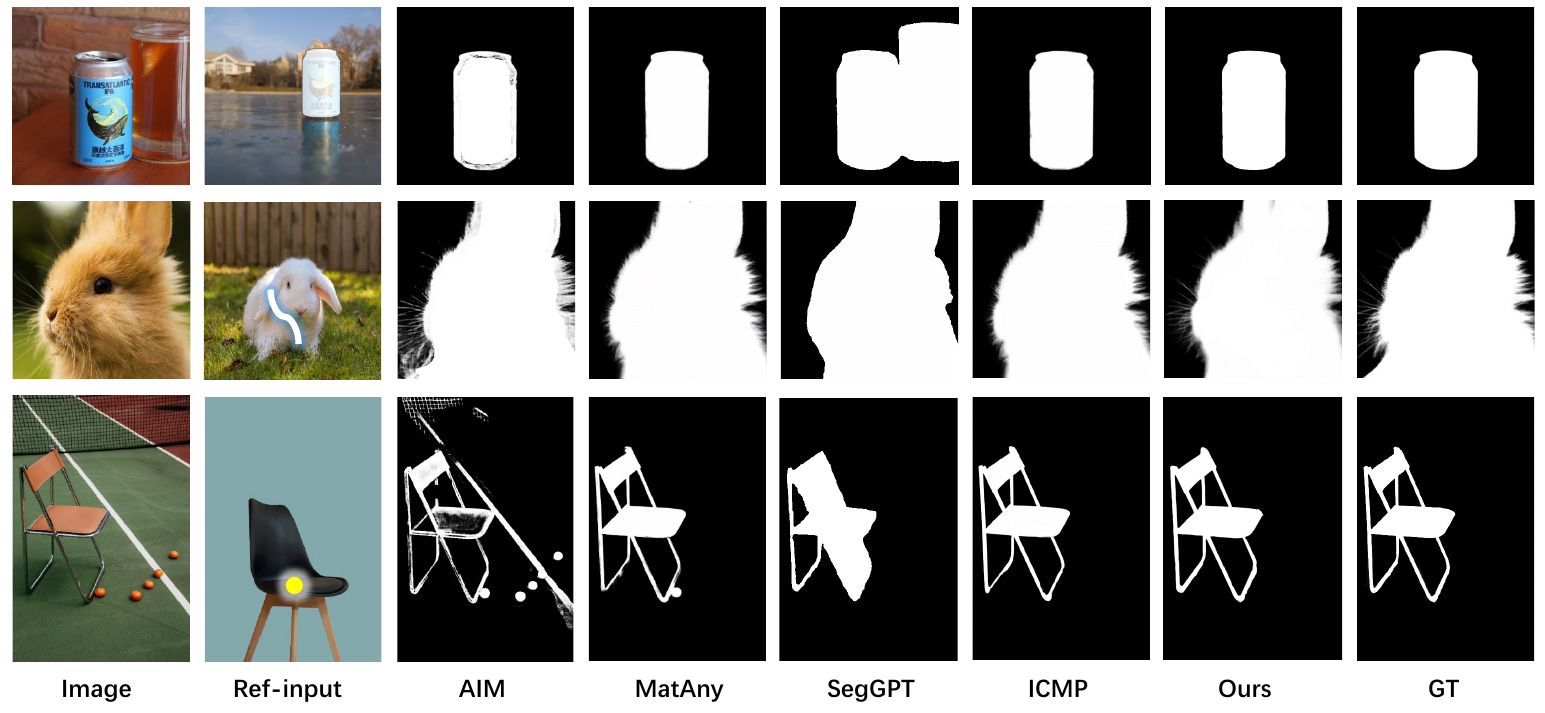}
  \vspace{-20pt}
  \caption{\textbf{Qualitative results of different image matting methods.} Our method can predict the alpha matte of the matting target specified by the reference input, offering notable prediction accuracy while avoiding interference from unrelated foreground elements.}
  \label{fig:com}
  \vspace{-5pt}
\end{figure*}


\paragraph{Comparison with In-Context Segmentation Models.}

To explore the effectiveness and superiority of our model, we selected SegGPT~\cite{wang2023seggpt} and SEEM~\cite{zou2023segment}, which also operates under in-context learning, as baselines in the image segmentation domain. From Table~\ref{tab:com_seg}, on the ICM-$57$ and AIM-$500$ datasets, under the same experimental setup (one mask per group of images), 
our model significantly outperforms the baselines across all metrics. 

It is important to note that due to their distinct task setting (\ie, segmentation rather than matting), the lack of high-quality annotations (\eg, alpha mattes) and the methods (\ie, mask classification rather than alpha matte regression), most of these models have overlooked pixel-level text-semantic alignment and are unable to produce fine-grained masks, as illustrated in Fig.~\ref{fig:com}.

\begin{table}[!t]
\footnotesize
\centering
\renewcommand{\arraystretch}{1}
\addtolength{\tabcolsep}{-4pt}
\begin{tabular}{@{}l|cccc|cccc@{}}
\toprule
\multicolumn{1}{c|}{\multirow{2}{*}{Method}} & \multicolumn{4}{c|}{ICM-$57$}     & \multicolumn{4}{c}{AIM} \\
\multicolumn{1}{c|}{}                        & MSE    & SAD   & GRAD  & CONN  & MSE & SAD & GRAD & CONN \\ \cmidrule(lr){1-1}\cmidrule(lr){2-5} \cmidrule(lr){6-9}
SegGPT                                       &0.0198 & 38.81 & 28.61 & 18.61 & 0.0391 & 42.65  & 41.95 & 26.69 \\
SEEM                                        &0.0292 & 64.28 & 37.54 & 23.64 & 0.0425 & 114.23 & 74.51 & 74.32 \\
Ours                                         &0.0081 & 19.12 & 18.65 & 11.21 & 0.0062 & 18.65  & 15.51 & 10.98 \\ \bottomrule
\end{tabular}

\vspace{-5pt}
\caption{\textbf{Comparison with in-context segmentation models.}}
\label{tab:com_seg}
\vspace{-5pt}
\end{table}

\vspace{-10pt}
\paragraph{Comparison with Automatic and Auxiliary Input-Based Matting Models.}
We further compare the performance of our IconMatting with both automatic and auxiliary input-based matting models on the ICM-$57$ and AIM-500 datasets. Automatic matting methods such as LF~\cite{zhang2019late} and AIM~\cite{AIM} lack specific auxiliary information about the matting target, often produce poor alpha mattes, 
showing a significant performance gap compared with our model. MGM~\cite{MGM} and MGMiW~\cite{MGMwild} use a mask for each image as auxiliary input to specify the matting target. Although our method simplifies this by requiring only one mask per group of images, it still outperforms MGM in various metrics. VitMatte~\cite{vitmatte}, a trimap-based image matting method, necessitates manually annotating a trimap for the foreground of each image, making it the performance upper bound for our in-context matting. Nevertheless, the performance of IconMatting is on par with VitMatte, underscoring its efficacy and competitiveness in image matting.

\begin{table*}[!t]
\centering
\small
\renewcommand{\arraystretch}{1.1}
\addtolength{\tabcolsep}{3pt}
\begin{tabular}{@{}l|c|cccc|cccc@{}}
\toprule
\multirow{2}{*}{Method} & \multicolumn{1}{c|}{\multirow{2}{*}{Guidance}} & \multicolumn{4}{c|}{ICM-$57$}     & \multicolumn{4}{c}{AIM}        \\ 
                        & \multicolumn{1}{c|}{}                          & MSE    & SAD   & GRAD  & CONN  & MSE    & SAD   & GRAD  & CONN  \\ \cmidrule(lr){1-1}\cmidrule(lr){2-2} \cmidrule(lr){3-6} \cmidrule(lr){7-10}
MGM                     & 1 mask per image                               & 0.0341 & 84.25  & 61.84 & 30.21  & 0.0268 & 71.91  & 23.37 & 21.97  \\
VitMatte                & 1 trimap per image                             & 0.0030 & 16.16  & 14.28 & 11.14  & 0.0038 & 18.79  & 14.22 & 12.47  \\
MGMiW                   & 1 mask per image                               & --  & -- & -- & --           & 0.0030 & 16.72  & 14.68 & 12.02  \\ \cmidrule(lr){1-1}\cmidrule(lr){2-2} \cmidrule(lr){3-6} \cmidrule(lr){7-10}
LF                      & auto                                           & 0.0811 & 205.68 & 69.53 & 195.63 & 0.0667 & 191.74 & 64.51 & 181.26 \\
AIM                     & auto                                           & 0.0265 & 65.07  & 55.56 & 25.74  & 0.0183 & 48.09  & 47.58 & 21.74  \\ \cmidrule(lr){1-1}\cmidrule(lr){2-2} \cmidrule(lr){3-6} \cmidrule(lr){7-10}
Ours                    & 1 mask per group of images                     & 0.0081 & 19.12  & 18.65 & 11.21  & 0.0062 & 18.65  & 15.51 & 10.98  \\ \bottomrule
\end{tabular}
\vspace{-5pt}
\caption{\textbf{Comparison with automatic and auxiliary input-based matting models.}}
\label{tab:com_autoaux}
\end{table*}

\vspace{-10pt}
\paragraph{Comparison with Interactive Matting Models.}

Recently, with the advent of SAM, some researchers have designed interactive matting models, such as MatAny~\cite{MatAny} and MAM~\cite{MAM}, based on this generic image segmentation model. Inspired by this, we also designed an In-Context Matting Pipeline (ICMP) with three stages: correspondence, segmentation, and matting, serving as one of the baselines. ICMP is a combination of existing models, with details available in supplementary materials.

For MatAny and MAM, we compare them under three types of interactions: points, scribbles, and masks. On the ICM-$57$ testing dataset, both baselines receive interaction information for each image within a group, whereas IconMatting only receive interaction information from one image in the group to indicate the matting target. Despite reducing the amount of human interaction, IconMatting achieves slightly better performance over the baselines.

Limited by the interaction modalities in ICMP, our model is only compared with it under the point interaction setting. Our end-to-end model outperforms the combined ICMP. In ICMP, the cues for the matting target degrade to points during the correspondence phase of the pipeline, often resulting in sparse information, making our end-to-end approach more effective.

\begin{table}[!t]
\centering
\footnotesize
\renewcommand{\arraystretch}{1.1}
\addtolength{\tabcolsep}{-3pt}
\begin{tabular}{@{}l|c|cccccc}
\toprule
\multicolumn{1}{c|}{\multirow{2}{*}{Method}} & \multirow{2}{*}{In-context} & \multicolumn{2}{c}{Point}& \multicolumn{2}{c}{Scribble}& \multicolumn{2}{c}{Mask}\\
\multicolumn{1}{c|}{}                        &                             & MSE    & SAD   & MSE   & SAD   & MSE    & SAD   \\ \cmidrule(lr){1-1}\cmidrule(lr){2-2} \cmidrule(lr){3-4} \cmidrule(lr){5-6}\cmidrule(lr){7-8}
MatAny                                       & \ding{55} & 0.0651 & 129.67 & 0.0512 & 115.21 & 0.0412 & 95.26 \\
MAM                                          & \ding{55} & 0.0149 & 41.23  & 0.0141 & 40.23  & 0.0109 & 29.65 \\
ICMP                                         & \ding{51} & 0.0112 & 39.65  & --  & -- & --  & -- \\
Ours*                           & \ding{55} & 0.0061 & 15.28  & 0.0059 & 15.97  & 0.0029 & 15.28 \\
Ours                            & \ding{51} & 0.0124 & 23.21  & 0.0105 & 24.56  & 0.0081 & 19.12 \\ 
\bottomrule
\end{tabular}
\vspace{-5pt}
\caption{\textbf{Comparison with interactive matting models.} In the penultimate row, our method is provided with guidance information for every image, reducing to an auxiliary input-based method. Our method outperforms automatic methods and some of the auxiliary input-based methods, and its performance is comparable to that of the trimap-based method, VitMatte.}
\label{tab:com_inter}
\end{table}

\subsection{Ablation Study}

\paragraph{Different Modules.}
To validate different modules, we conducted ablation studies in Table~\ref{tab:ablation}. Among the four components, the presence or absence of inter- and intra-similarity plays a crucial role in performance. Without intra-similarity, the performance of the model significantly worsens across all four metrics. If both inter- and intra-similarity are absent, the model degenerates to directly predicting the alpha matte from the image, losing the information source for the specified matting target, and thus the performance markedly deteriorates.

\vspace{-10pt}
\paragraph{Number of Reference Inputs.}
Intuitively, the more reference inputs there are, the more likely the model is to identify the corresponding matting target. We explored the impact of the number of reference inputs on the performance of our model, as shown in Table.~\ref{tab:ref_num}. On ICM-$57$ test set, the performance improves as the number of reference inputs increases; however, the improvement almost ceases when the number of reference inputs increases from 3 to 4. Therefore, we can conclude that appropriately increasing the number of reference inputs can enhance model performance, which is consistent with intuition.




\begin{table}[!t]
\centering
\footnotesize
\renewcommand{\arraystretch}{1}
\addtolength{\tabcolsep}{0pt}
\begin{tabular}{@{}cccccccc@{}}
\toprule
INTER & INTRA & MF  & SD  & MSE    & SAD   & GRAD  & CONN  \\ 
\midrule
\ding{51}   & \ding{51}   & \ding{51} & \ding{51} & 0.0081 & 19.12 & 18.65 & 11.21 \\
\ding{51}   & \ding{55}   & \ding{51} & \ding{51} & 0.0099 & 24.15 & 20.36 & 12.11 \\
\ding{55}   & \ding{55}   & \ding{51} & \ding{51} & 0.0315 & 40.32 & 31.56 & 19.53 \\
\ding{51}   & \ding{51}   & \ding{55} & \ding{51} & 0.0054 & 23.69 & 21.96 & 14.52 \\
\ding{51}   & \ding{51}   & \ding{51} & \ding{55} & 0.0071 & 18.56 & 19.54 & 12.61   \\ 
\bottomrule
\end{tabular}
\vspace{-5pt}
\caption{\textbf{Ablation study on different modules.} INTER, INTRA, MF, and SD respectively stand for inter-similarity, intra-similarity, multi-scale features, and segmentation dataset.}
\label{tab:ablation}
\end{table}

\begin{table}[!t]
\footnotesize
\renewcommand{\arraystretch}{1}
\addtolength{\tabcolsep}{5pt}
\centering
\begin{tabular}{@{}ccccc@{}}
\hline
\#Reference & MSE    & SAD   & GRAD  & CONN  \\ \hline
1                  & 0.0085	& 19.58	& 19.14	& 12.61  \\
2                 & 0.0075 & 16.57 & 18.52 & 11.15   \\
3                 & 0.0070 & 15.48 & 17.56 & 10.56 \\
4                 & 0.0068 & 15.23 & 17.02 & 10.28 \\ \hline
\end{tabular}
\vspace{-5pt}
\caption{\textbf{Ablation study on the number of reference inputs.}}
\label{tab:ref_num}
\end{table}

\subsection{Extension to Video Object Matting}
\begin{figure}[!t]
  \centering
  \includegraphics[width=0.95\linewidth]{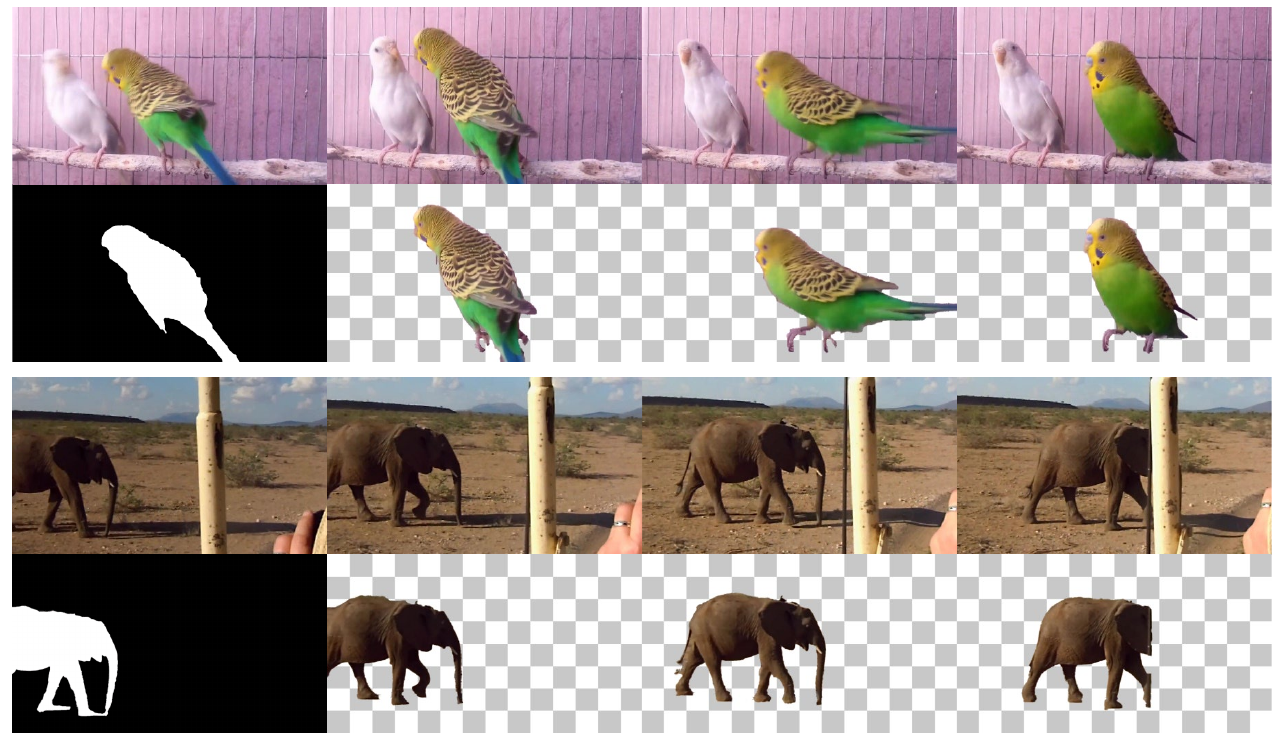}
  \vspace{-5pt}
  \caption{\textbf{Extension to video object matting.}}
  \label{fig:video}
  \vspace{-5pt}
\end{figure}

The technique of in-context matting is easily extendable to video object matting. The key is to use a frame of the video as a reference. For example, an object is marked in the first frame of a video, which serves as a reference input and all frames of the video are treated as target images. With this setup, the model for in-context matting can predict the alpha matte for each frame of the video, visualized in Fig.~\ref{fig:video}.

\section{Conclusion}

In this work, we introduce 
`in-context matting', which enables automatic matting of foreground of interest on target images given a reference image and its prompt. We introduce IconMatting as a preliminary solution. Extensive experiments have shown its efficacy and robustness across categories and scenes. Being the first work introducing this task, we believe it opens new possibilities for efficient and accurate 
image matting while reducing user effort, also enhancing the versatility of image matting techniques.  

\vspace{5pt}
\noindent\textbf{Acknowledgement.} This work is supported by the National Natural Science Foundation of China under Grant No. 62106080.

{
    \small
    \bibliographystyle{ieeenat_fullname}
    \bibliography{main}
}

\maketitle
\clearpage
\setcounter{page}{1}
\newpage
{
    \centering
    \Large
    \textbf{\thetitle}\\
    \vspace{0.5em}Supplementary Material \\
    \vspace{1.0em}
}

\begin{figure*}[htb]
  \centering
  \includegraphics[width=\linewidth]{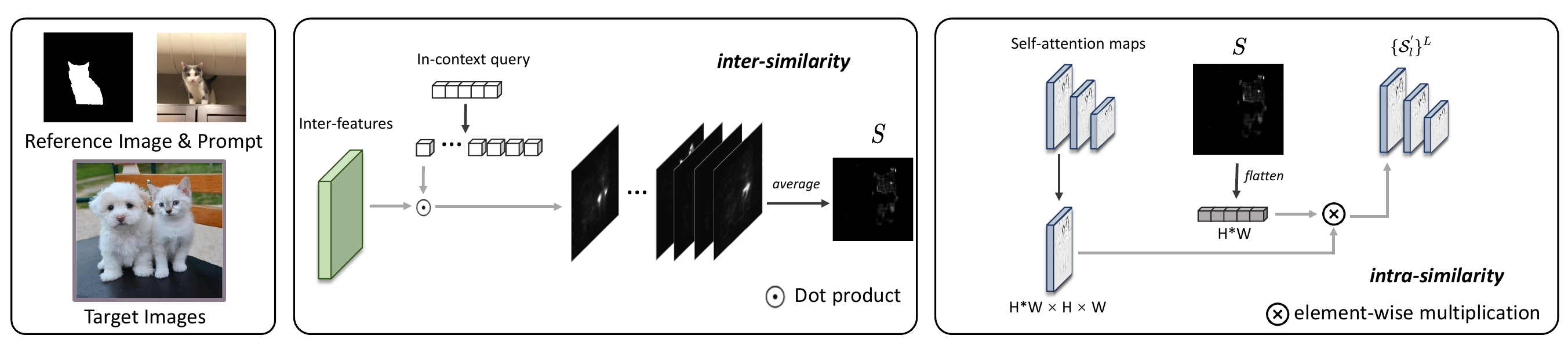}
  \caption{\textbf{Illustration of the inter- and intra-similarity modules.} For simplicity, the resize operation is omitted, only the calculation of one element of the in-context query is depicted, and the fusion process of self-attention maps from a single scale is shown.}
  \label{fig:inter-intra}
\end{figure*}

\section{Overview}

The supplementary material includes the following sections:

\begin{itemize}

\item Implementation details of IconMatting. 
\item Additional experiments.
\item Dataset.
\item Implementation of a baseline-In-Context Pipeline.

\end{itemize}  

\section{Implementation Details of IconMatting}

Here, we supplement schematic diagrams for the inter- and intra-similarity modules.
The inter-similarity computes the similarity between features extracted from the target image and the in-context query derived from the reference image, generating an average similarity map, $S$. The intra-similarity combines the self-attention maps representing intra-image similarities within the target image with the similarity map $S$ obtained from the inter-similarity module. This fusion employs elements of $S$ as weights assigned to the self-attention maps, thereby providing guidance information for the matting target.




\section{Additional Experiments}
\subsection{More Qualitative Results}
\begin{figure*}[htbp]
  \centering
  \includegraphics[width=\linewidth]{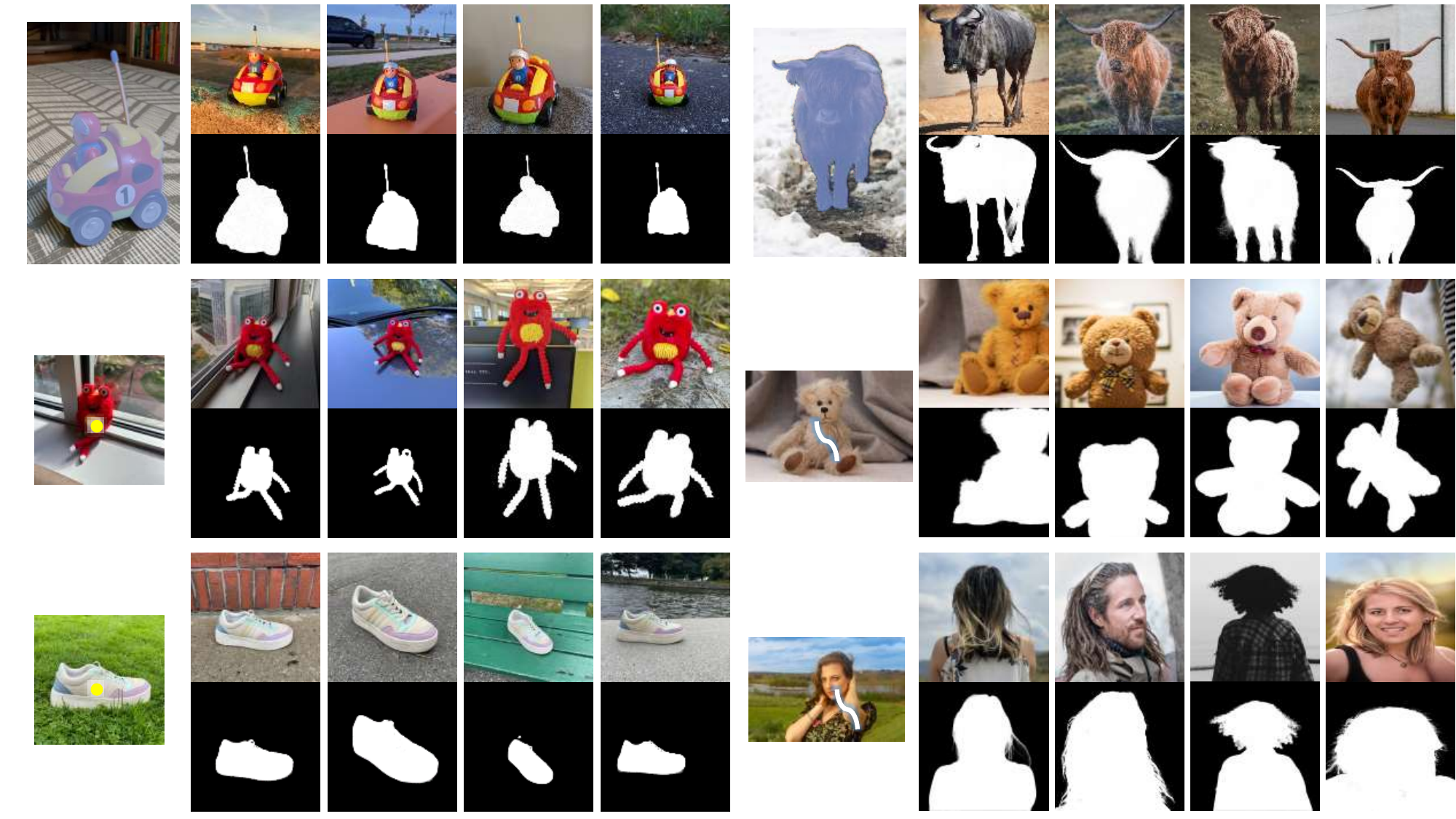}\vspace{5pt}
  \caption{\textbf{Qualitative results of IconMatting.} The first column shows the reference input, while the remaining columns display target images and their predicted alpha mattes. Given a single reference input, our method can automatically process the same instance or category of foreground.}
  \label{fig:icm}
\end{figure*}

Here, we visualize more qualitative results of IconMatting in real-world scenarios, as shown in Fig.~\ref{fig:icm}.
\vspace{5pt}
\subsection{Why Composited Dataset is Not Used?}

Our IconMatting, developed as a model for image matting, underwent exclusive training solely on real-world datasets, omitting composited data—a practice uncommon within the domain of image matting. This methodology was adopted due to the substantial discrepancy observed in model performance between composited and real-world datasets when employing Stable Diffusion as the backbone. To substantiate this assertion, we conducted experiments on Composition-1k and RM-1k datasets. For IconMatting, we modified it to be a trimap-based image matting model. First, we removed the in-context similarity module of IconMatting. Second, we concatenated the trimap with the original image and feed them together into the matting head. The training and test sets of Composition-1k were pre-defined, while RM-1k underwent training and test set partitioning in an 8:2 ratio.

The experimental findings, as depicted in Tab.~\ref{tab:cr}, reveal a significant performance gap between RM-1k and Composition-1k datasets. Despite the considerably smaller sample size in RM-1k compared to Composition-1k, the former demonstrated notably superior test results. This discrepancy highlights that, Stable Diffusion, pre-trained for generative tasks, does not perform optimally for image matting tasks on composited datasets within the context of our study.

\begin{table}[htbp]
\renewcommand{\arraystretch}{1}
\addtolength{\tabcolsep}{-1pt}
\centering
\begin{tabular}{lcccc}
\toprule
\multicolumn{1}{c}{Dataset} & MSE                  & SAD                  & GRAD                 & CONN                 \\ \midrule
Composited   dataset        & 0.0227               & 110.12               & 68.84                & 59.02                \\
Real-world   dataset        & 0.0029               & 14.11                & 16.32                & 13.56                \\ \bottomrule
                            & \multicolumn{1}{l}{} & \multicolumn{1}{l}{} & \multicolumn{1}{l}{} & \multicolumn{1}{l}{}
\end{tabular}
\caption{\textbf{Experiment on composited dataset and real-world dataset.}}
\label{tab:cr}
\end{table}

\subsection{Ablation on Diffusion Time Steps}

When using Stable Diffusion as a feature extractor, the choice of the time step $t$ during the diffusion process is crucial. Generally, a small $t$ corresponds to features that represent the image more closely, with minimal noise added, which is why methods like ODISE~\cite{xu2023open} for image segmentation use $t=0$. Conversely, a large $t$ captures more abstract semantic features, as seen in DIFT~\cite{tang2023emergent} for semantic correspondence using $t=261$.

For IconMatting, we aim to extract features that both express abstract semantics 
in the inter-similarity module and retain detailed characteristics of the image for predicting the alpha matte with the matting head. Therefore, selecting the appropriate $t$ is a 

trade-off. As shown in Table~\ref{tab:timestep}, our experiments show that performance is suboptimal for rather small $t$ values (\eg, $0$) or too large $t$ values (\eg, over $300$). The optimal performance is achieved when $t$ is set to $200$.

\begin{table}[htbp]
\renewcommand{\arraystretch}{1}
\addtolength{\tabcolsep}{6pt}
\centering
\begin{tabular}{@{}ccccc@{}}
\toprule
Timesteps  & MSE    & SAD   & GRAD  & CONN  \\ 
\midrule
0   & 0.0091 & 22.12 & 20.18 & 10.98 \\
100 & 0.0094 & 21.01 & 19.54 & 10.94 \\
200 & 0.0081 & 19.12 & 18.65 & 11.21 \\
300 & 0.089  & 23.84 & 20.11 & 12.46 \\
400 & 0.098  & 31.39 & 24.57 & 14.28 \\ 
\bottomrule
\end{tabular}
\vspace{5pt}
\caption{\textbf{Ablation study on the choice of diffusion timesteps.}}
\label{tab:timestep}
\end{table}

\subsection{Visualization on In-Context Similarity}
To further illustrate the effectiveness of our core module, the in-context similarity, we visualize both inter- and intra-similarity modules. For inter-similarity, we directly visualize \( S \); for intra-similarity, we resize multi-scale \( \{ {\cal S}_l^{'} \}^{L} \) to a uniform scale, averaged it, and then visualized the result. This is demonstrated in the Fig.~\ref{fig:sim_vis}.

In the case of an alarm clock and a monkey, due to some differences between the reference input and the matting target, the lower left part of the matting target is lost in the results of inter-similarity. However, by relying on the intra-similarity, the results of intra-similarity complement and complete the matting target, thus predicting a complete alpha matte.

This analysis underscores the significance of considering both inter- and intra-similarity in our approach. 
\begin{figure*}[htbp]
  \centering
  \includegraphics[width=\linewidth]{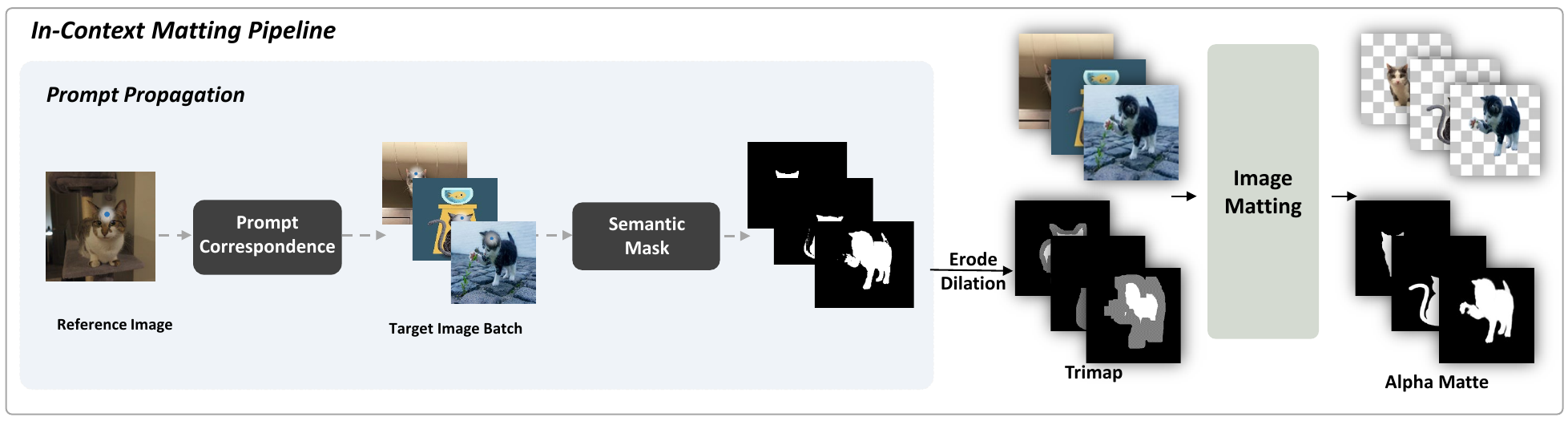}
  \caption{\textbf{In-Context Matting Pipeline (ICMP)} consists of two parts: prompt propagation module and interactive image matting module. Prompt propagation module generates prompts for target images based on the prompts on reference image through semantic correspondence, then interactive image matting module predicts alpha matte with images and prompts.}
  \label{fig:pipeline_}
\end{figure*}

\begin{figure}[htbp]
  \centering
  \includegraphics[width=\linewidth]{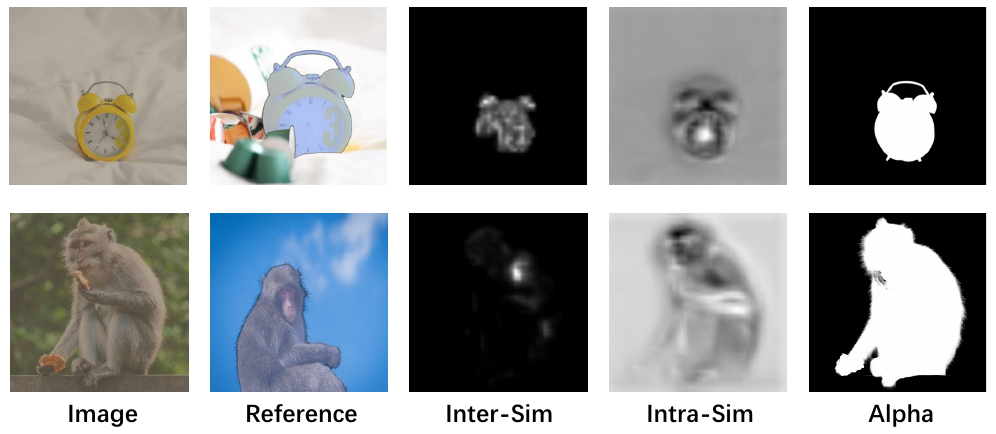}
  \caption{\textbf{Visualization on in-context similarity.}}
  \label{fig:sim_vis}
\end{figure}

\section{Dataset}
\begin{figure}[htbp]
  \centering
  \includegraphics[width=\linewidth]{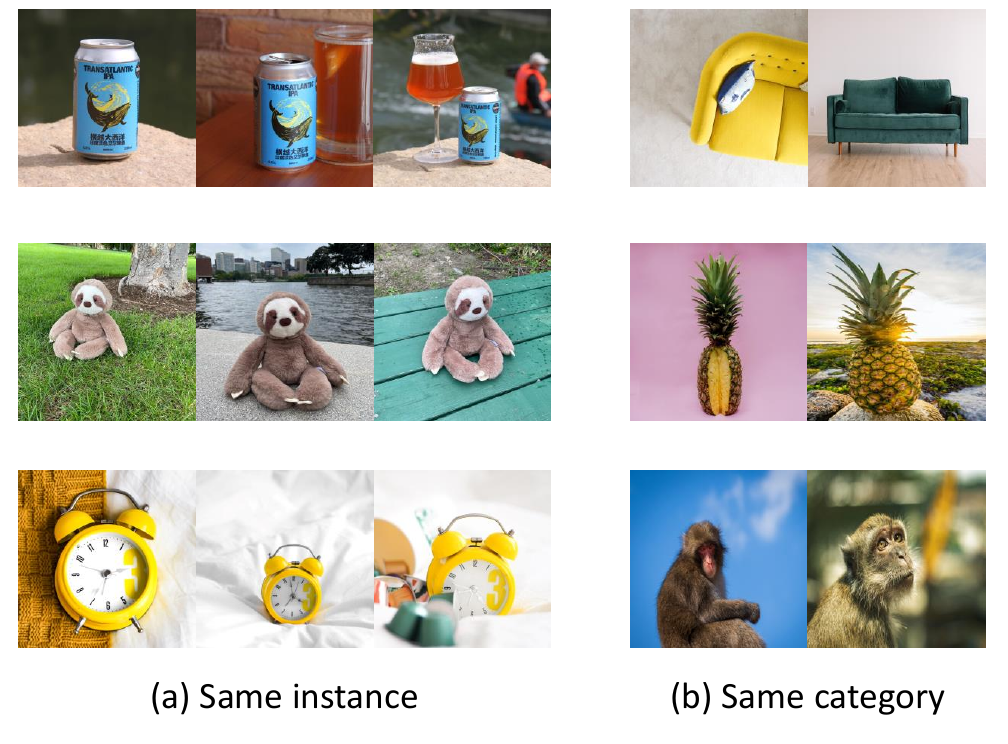}
  \caption{\textbf{ICM-$57$} encompasses foregrounds of both the same category and same instance.}
  \label{fig:data_supp}
\end{figure}

\begin{figure}[htbp]
  \centering
  \includegraphics[width=\linewidth]{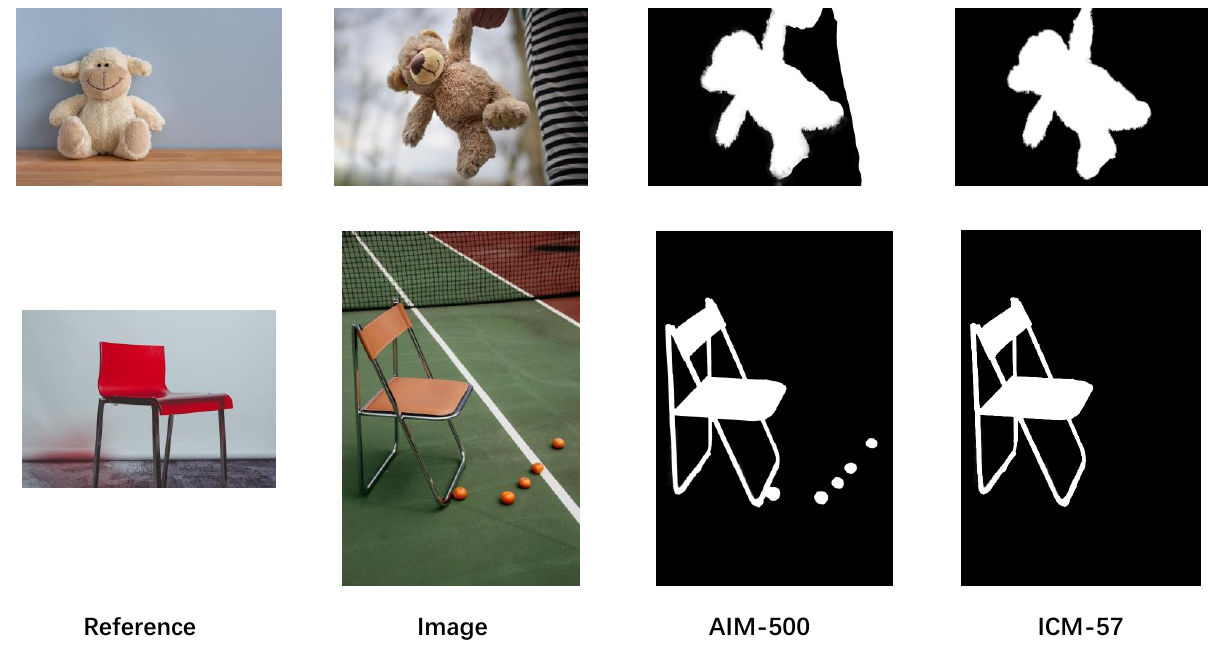}
  \caption{\textbf{Modifications made to the annotations.} In the second row of pictures, only the chair is preserved in the alpha matte, which meets the needs of in-context matting.}
  \label{fig:modify}
\end{figure}

Here, we show more samples of our dataset and add details about the construction of our dataset.

For our test set ICM-$57$, as described in the main text, it encompasses foregrounds of the same category and same instance, fulfilling the essence of in-context matting. Examples of such instances are depicted in Fig~\ref{fig:data_supp}. In order to utilize a portion of images from AIM-$500$, we modified their annotations to align with the requirements of in-context matting, as illustrated in Fig.~\ref{fig:modify}.

\section{In-context Matting Pipeline}

In our novel task setting for image matting, there wasn't an existing baseline available for direct comparison with our IconMatting. Therefore, leveraging some off-the-shelf  models, we established a pipeline aimed at achieving in-context matting, serving as a baseline in the experiment in the main text.

In-Context Matting Pipeline (ICMP) consists of two modules: a prompt propagation module and an image matting module, as shown in Fig. \ref{fig:pipeline_}. 
With ICMP, users provide points as prompts on reference images to indicate matting targets, then the prompt propagation module extends these prompts to all input images and the semantic masks of the corresponding matting targets are obtained. Subsequently, the image matting module processes the input images and corresponding semantic masks, generating alpha mattes for the specified matting targets across all images.

We utilize the semantic correspondence property of features provided by DINOv2~\cite{oquab2023dinov2} to achieve prompt correspondence, and use SAM~\cite{kirillov2023segment} to extract coarse semantic masks corresponding to the prompt points, thus realizing our designed prompt propagation, and thus ICMP.

\subsection{Prompt Propagation Module}

 While auxiliary input-based image matting can yield the alpha matte for a user-specified matting target, it requires manual prompting for each input image, even when the matting target is the same. Prompt propagation addresses this issue by disseminating the prompts provided on the example image to all input images, resulting in a set of semantic masks. 
 

In essence, prompt propagation can be likened to semantic correspondence, wherein prompts from the example image are matched to corresponding prompts in other images. By prompt propagation, the user's prompt for the example image can be propagated to the other images, eliminating the need to manually provide a prompt for each image.

Considering that features extracted by DINOv2, a model pretrained on large-scale datasets, exhibit strong generalization to real-world data for semantic correspondence without further training, we employ DINOv2 to propagate prompt points for the example image to other images. Given an input image and prompt points indicating the matting target, SAM can be used to generate a semantic mask corresponding to the prompt.


\subsection{Image Matting Module}
This module enables the extraction of the alpha matte for the matting target based on semantic masks from prompt propagation. 

By applying morphological operations to the mask, we transform a semantic mask from prompt propagation into a pseudo-trimap. Then, any trimap-based image matting model can be used to obtain a alpha matte for the matting target, and our choice is VitMatte~\cite{vitmatte}.



\end{document}